# High Level Path Planning with Uncertainty

**Runping Qi    David Poole**
Department of Computer Science
University of British Columbia
Vancouver B. C. Canada V6T 1W5
E-mail: qi@cs.ubc.ca, poole@cs.ubc.ca

## Abstract

For high level path planning, environments are usually modeled as distance graphs, and path planning problems are reduced to computing the shortest path in distance graphs. One major drawback of this modeling is the inability to model uncertainties, which are often encountered in practice. In this paper, a new tool, called *U-graph*, is proposed for environment modeling. A U-graph is an extension of distance graphs with the ability to handle a kind of uncertainty. By modeling an uncertain environment as a U-graph, and a navigation problem as a Markovian decision process, we can precisely define a new optimality criterion for navigation plans, and more importantly, we can come up with a general algorithm for computing optimal plans for navigation tasks.

## 1 INTRODUCTION

For high level path planning, digraphs (distance graphs) are usually used as a tool for environment modeling. In a digraph, vertices denote places (or landmarks), edges — vertex pairs — denote the routes between the place pairs, and the weight of an edge denotes the cost (or distance) between the two vertices of the edge. The path planning problem is formulated as a *shortest path* problem in digraphs.

However, a major drawback of digraphs is the inability to model uncertainties. If there is an edge between two vertices in a digraph, it is assumed that there is a direct route between the places represented by these vertices. But, in reality, we often encounter some situations where we are not sure about something. For example, we may know that there is a door between two rooms, but that door may be locked; we are not sure whether the door is open now, though we know that, according to past experience, the probability of the door being open is about 0.8. Clearly, digraphs are not sufficient to model such situations.

As the first effort to extending the expressive power of digraphs, we propose in this paper a new kind of graphs, called *U-graphs* (uncertain graphs), and investigate path planning problems based on the U-graph model. A U-graph is an ordinary digraph augmented with a new kind of edges: *switches*. Like an edge, each switch connects a vertex pair and has a weight. In addition, each switch has a probability associated with it. The probability associated with a switch represents the probability that the connection between the two vertices of the switch is traversable.

With uncertainty being taken into consideration, the path planning problem is significantly different from the shortest path problem. Instead, a path planner needs to compute a "navigation plan" which can result in an optimal travel with respect to some predefined measures. The main difficulty in computing such an optimal travel strategy largely attributes to the *on-line* nature of the problem, that is, an optimal strategy should be able to tell an agent what is the optimal next step, based on the incomplete knowledge known so far, in any possibly encountered situation.

Traditionally, the quality of an on-line algorithm is measured based on two criteria [Bar-Noy and Schieber, 1991]: *competitive ratio* of the algorithm and worst-case performance. In this paper, we adopt a new optimality criterion in terms of the expected cost for a given task, which we think is very suitable for navigation. We formulate a navigation problem as a *finite state Markovian decision process*. Markovian decision process was studied as a mathematical abstraction of certain types of dynamic systems [Derman, 1970] and as a branch of dynamic programming [Howard, 60] [Denardo, 1982]. With this formulation, the path planning problem amounts to the *optimal first-passage problem* for a Markovian decision process [Derman, 1970]. From this formulation, we derive an algorithm for path planning.

The rest of this paper is organized as follows. In the next section, U-graphs are formally introduced and world modeling based on U-graphs is briefly discussed. In Section 3, path planning in uncertain environments is addressed and an optimality criterion for navigation plans is informally presented. In Section 4, the path planning problem is formalized and a general path planning algorithm is derived. Section 5 discusses related work, and Section 6 concludes the paper with a brief discussion of our future work.

## 2  U-GRAPH

**Definition 1** *A* U-graph *is* $\langle V, E, S_u, pr, weight \rangle$, *where* $V$ *is a finite set of vertices,* $E \subseteq V \times V$ *is a set of edges,* $S_u \subseteq V \times V$ *is a set of switches, pr a probability function from* $S_u$ *to* $[0, 1]$, *and weight is a function from* $E \cup S_u$ *to* $R^+$.

For the purpose of clarity and simplicity, we assume that $E$ and $S_u$ are mutually disjoint.

For a U-graph $\langle V, E, S_u, pr, weight \rangle$, its *pessimistic induced graph* is graph $\langle V, E \rangle$; its *optimistic induced graph* is graph $\langle V, E \cup S_u \rangle$. A U-graph is said to be *disconnected wrt* two vertices if there is no path between the two vertices in the optimistic induced graph.

For high level path planning, the essential knowledge a path planner needs to know about the environment is the topology and connectivity of the environment. The topological structure of an environment can be represented by a set of "interesting places" and the connectivity relationships among these places. These places can be abstracted as vertices and the connectivity relationships abstracted as arcs in a U-graph. If it is uncertain whether the route between the two places is definitely traversable, this uncertainty can be abstracted by a switch. We say a switch is *on* if it is traversable, and a switch is *off* if it is not traversable. The probability that the route being traversable is assigned to the switch. The weight of the switch denotes the cost for an agent to go through the switch provided that it is on.

As an illustration of how to model a practical situation by a U-graph, let us consider a situation as shown in Figure 1-(a), where two places A and B are on the two sides of a river, and usually connected by a bridge. However, it is not sure whether the bridge is broken or not, though it is known that the probability for the bridge being broken is 0.2. If the bridge is not broken, it will take about five minutes to cross the river. The connection between A and B can be represented as a switch, as shown in Figure 1-(b).

In the following discussion we make three assumptions about a U-graph. First, the probabilities of the switches in $S_u$ are mutually independent. Second, the status of the switches in $S_u$ is uncertain to the agent unless the agent is at either end vertex of the switch.

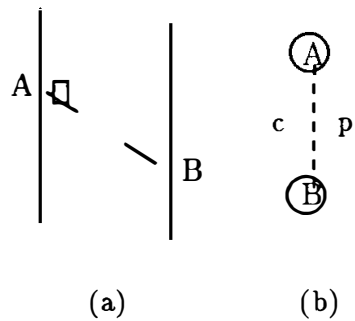

(a)                    (b)

Figure 1: Modeling an uncertain situation by a U-graph

When an agent is at one end of a switch $s$, the agent can reach the other end through the switch with the cost given by $weight(s)$ if the switch is actually on, and cannot otherwise traverse the switch. Third, the status of any switch will not change after the agent discovers it. The first two assumptions are justifiable in most situations. The third assumption is also justifiable because a U-graph is mainly used to represent the overall structure of an environment, and it is reasonable to believe that changes in the overall structure of an environment are comparatively slow.

The third assumption above implies that, when the agent reaches one ending vertex of a switch in $S_u$, the switch can be replaced by an edge with the same weight if the status of the switch turns out to be on, and can be deleted from the graph otherwise.

## 3  PATH PLANNING WITH UNCERTAINTY

As a motivating example, let us consider the case shown in Figure 2. Suppose that an agent is at vertex A and is asked to go to vertex B. In order to accomplish this task, the first question to be answered is: "which route should be taken?". Here are some possible answers.

First, if the agent wants to minimize the cost in the worst case, it should take the upper route (edge AB) in the graph.

Second, if the agent would like to minimize the expected cost for accomplishing this task, there are two choices. The first one is to go to B through edge AB. Another one is as follows: go to C first; if CD is actually traversable, go to D then to B; otherwise go back to A, then go to B through edge AB. The expected cost for the first choice is $d_1$. The expected cost for the second choice is:

$$d_2 + p * (d_4 + d_3) + (1 - p) * (d_2 + d_1)$$

Third, if the agent has a resource limit, it may want to choose a navigation plan which can maximize the probability of reaching the destination within the resource limit.

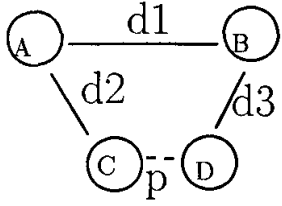

Figure 2: A simple path planning case with uncertainty

As we see from the above example, an agent may be subject to various constraints and want to achieve various objectives for a given navigation task. These constraints and objectives determine the optimality criteria of plans. In this paper, we consider path planning problem *wrt* the criterion of minimization of the expected cost for achieving a given task.

More precisely, we state the navigation problem and the optimality criterion as follows. *When given a task of going to vertex q from vertex p in a U-graph representation of an environment, an agent is supposed to systematically explore the environment until either arriving at the goal position or finding out that there is actually no path to the goal position. The path planning problem is to determine a navigation plan which minimizes the expected cost required for exploring the environment.*

## 4  A FORMAL DEFINITION OF PATH PLANNING

In the present section, we model a navigation problem as a *Markovian decision process* [Derman, 1970], a navigation plan as a *policy* for the Markovian decision process, and give a precise definition of the expected cost for a navigation plan.

### 4.1  Markovian Decision Model

Informally, a Markovian decision process is an alternating sequence of states of, and actions on, an evolution system. At each point of time, the state of the system can be observed and classified, and an action, based on the observed state, can be taken. A *policy* is a prescription for taking action at each point in time.

Formally, a Markovian decision process is a quadruple $\langle I, K, w, q \rangle$, where $I$ denotes the space of the states that can be observed of a system, $K = \{K_i | i \in I\}$, $K_i$ denotes the set of actions which may be taken in state $i$, $w$ is a cost function and $q$ is a transition function.

The laws of motion of the system are characterized by the transition function $q$. Whenever the system is in state $i$ and action $a$ is taken, then, regardless of its history, $q(i, j, a)$ denotes the probability of the system being in state $j$ at the next instant the system is observed. It is assumed that $\sum_j q(i, j, a) = 1$ for any $i \in I$ and $a \in K_i$. A cost structure is superimposed on a Markovian decision process. Whenever the system is in state $i$ and action $a$ is taken, a known cost $w(i, a)$ is incurred.

A deterministic policy for a Markovian decision process can be thought as a function mapping from states to actions. For the purpose of this paper, we will only consider deterministic policies. For a Markovian decision process and a fixed policy $R$, let $P_R\{Y_t = i\}$ be the probability of $M$ being in state $i$ at time $t$ under the control of policy $R$. We define a set of random variables $\{W_t, t = 0, 1, ...\}$:

$$W_t = w(i, R(i)) \quad \text{if } Y_t = i.$$

The expected cost at time $t$ *wrt* policy $R$ is:

$$E_R\{W_t\} = \sum_i P_R\{Y_t = i\} w(i, R(i)).$$

Let $Y_0 = i$ be the initial state and let

$$S_{R,T}(i) = E_R \sum_{t=0}^{T} W_t = \sum_{t=0}^{T} \sum_j P_R\{Y_t = j\} w(j, R(j)).$$

$S_{R,T}(i)$ is the expected total cost of operating the system up to and including the time "horizon" $t = T$, given the initial state $i$ and policy $R$. The *optimal first-passage problem* is to find $R$ that minimizes $\lambda_R(i) = S_{R,\tau}(i)$, where $\tau$ denotes the smallest positive value of $t$ such that $Y_t = j$, and $j$ is one of the target states at which the process is stopped. This problem is was first formulated by Eaton and Zadeh [Eaton and Zadeh, 1962].

Let $C_D$ denote a class of all (deterministic) policies. Derman [Derman, 1970] proved that: if $\{w(i, a)\}$ are non-negative, then there exists an $R* \in C_D$ such that

$$\lambda_{R*}(i) = \inf_{R \in C_D} \lambda_R(i), \quad i \in I.$$

For a given policy $R$, let $X(R, i)$ denote the expected value of the total cost of reaching the target state from state $i$ and $R*$ denote the optimal policy. Thus, we have:

$$X(R*, i) \leq X(R, i) \tag{1}$$

and

$$X(R, i) = E\{w(i, a) + X(R, j)\} \tag{2}$$

or

$$X(R, i) = w(i, a) + \sum_{j \in I} q(i, j, a) * X(R, j) \tag{3}$$

for all $i \in I$, where $a = R(i)$. Furthermore, if there exists another policy $R'$ such that for some state $i$,

$X(R', i) < X(R, i)$, then $R$ must not be an optimal policy.

The behaviors of a Markovian decision process can be represented as a directed graph. Formally, the *representing graph of the Markovian decision process* $M = (I, K, w, q)$ is a directed graph $RG(M) = \langle V, A_1 \cup A_2 \rangle$ defined as:

$$V = I \cup \{s_{ia} | i \in I; a \in K_i\};$$
$$A_1 = \{\langle i, s_{ia}\rangle | i \in I; a \in K_i\}$$

and

$$A_2 = \{\langle s_{ia}, j\rangle | i, j \in I; a \in K_i; q(i, j, a) > 0\}$$

In such a representing graph, a node $i \in I$, called a *state node*, represents an observable state while node $s_{ia}$, called a *non-state node*, represents a temporary state resulting from taking action $a$ in state $i$. The next observable state after the temporary state is determined by a probability distribution $q(i, j, a)$. Therefore we can attach action $a$ as the label for arc $\langle i, s_{ia} \rangle$ and probability $q(i, j, a)$ as the label for arc $\langle s_{ia}, j \rangle$.

The *representing graph for a Markovian decision process $M$ starting with state $i_0$*, denoted by $RG_{i_0}(M)$, is the largest subgraph of $RG(M)$ such that any node in the subgraph is reachable from $i_0$.

In particular, the behaviors of a Markovian decision process controlled by a particular policy $R$ can be represented by a directed graph $RG(M, R)$ defined as:

$$V = I \cup \{s_{ia} | i \in I; a \in K_i\}$$

and

$$A_1 = \{\langle i, s_{ia}\rangle | i \in I; a \in K_i; R(i) = a\}$$

and $A_2$ is the same as the above. The representing graph for a Markovian decision process $M$ controlled by policy $R$ starting with state $i_0$, denoted by $RG_{i_0}(M, R)$, is the largest subgraph of $RG(M, R)$ such that any node in the subgraph is reachable from $i_0$ in the original graph.

Such a graph representation is quite useful. With this representation, we can study the properties of a Markovian decision process by studying its representing graph. More importantly, it can facilitate the derivation of the algorithm for computing the optimal plan for a given navigation task (see section 4.4).

### 4.2 Some Related Concepts

A *configuration* is a triple $\langle G, n_c, n_g \rangle$, where $G$ is a U-graph, $n_c$ and $n_g$ are two vertices of the graph, representing the current position and the goal position respectively. An edge is said to be a *current edge* of a configuration if the current vertex of the configuration is one of the vertices of the edge. Similarly, a switch is said to be a *current switch* of a configuration if the current vertex of the configuration is one of the vertices of the switch. Let $CE(C)$ and $CS(C)$ denote the set of current edges and the set of current switches of configuration $C$ respectively.

A *terminal* configuration (or a terminal for short) is a configuration which satisfies either of the following two conditions:

1. the shortest distance between $n_c$ and $n_g$ in the optimistic induced graph $\langle V, E \cup S_u \rangle$ is equal to the shortest distance between $n_c$ and $n_g$ in the pessimistic induced graph $\langle V, E \rangle$;
2. $G$ is disconnected *wrt* to $n_c$ and $n_g$.

In other words, a terminal is a configuration which is certain enough such that the agent can determine whether there is a path to the goal position, and be able to compute the optimal path if there is any. A terminal satisfying the first condition will be called a "good" terminal. A terminal satisfying the second condition will be called a "bad" terminal. A configuration $C$ is said to be an *uncontrolled configuration* if $CS(C)$ is not empty. A configuration is *controlled* if it is not uncontrolled.

Two kinds of *transitions*, namely *uncontrolled* and *controlled transitions*, can be defined respectively on uncontrolled and controlled configurations. For a controlled configuration $C = \langle G, n_c, n_g \rangle$, suppose $n_1, ..., n_k$ are the $k$ ($k > 0$) neighbours of $n_c$; we define $k$ controlled transitions, $t_1, ..., t_k$, corresponding to the $k$ current edges. We denote by $trans(C, t_i)$ the configuration to which the $i$-th transition $t_i$ can lead from configuration $C$. We have:

$$trans(C, t_i) = \langle G, n_i, n_g \rangle.$$

For an uncontrolled configuration $C = \langle G, n_c, n_g \rangle$, suppose $G = \langle V, E, S_u, pr, weight \rangle$ and $CS(C) = \{s_1, ..., s_k\}$ is the set of $k$ ($k > 0$) current switches of $C$; we define $2^k$ uncontrolled transitions corresponding to all the possible combinations of the $k$ switches' states. We denote by $ONS_{t_i}$ the set of switches in $CS(C)$ which are on, and by $OFFS_{t_i}$ the set of switches in $CS(C)$ which are off. We denote by $p_{t_i}$ the probability that the transition $t_i$ can happen. Thus,

$$p_{t_i} = \prod_{s \in ONS_{t_i}} pr(s) * \prod_{s \in OFFS_{t_i}} (1 - pr(s))$$

and

$$trans(C, t_i) = \langle G', n_c, n_g \rangle$$

where $G' = \langle V, E', S'_u, pr', weight \rangle$, where $E' = E \cup ONS_{t_i}$ and $S'_u = S_u - CS(C)$.

As an example, Figure 3-(a) shows a controlled configuration where the goal position is vertex $q$ and the current position is indicated by a double-circle. Three controlled transitions are associated with this configuration, corresponding to going to vertices 2, 3, and 4 respectively. Suppose the transition of going to vertex 4 is chosen, then the next configuration, as shown in Figure 3-(b), is an uncontrolled one. With this uncontrolled configuration, four uncontrolled transitions are associated, corresponding to the possible combinations of the states of switches $s_1$ and $s_2$. The possible configurations to which these transitions can lead are shown in Figures 3-(c), 3-(d), 3-(e) and 3-(f) respectively.

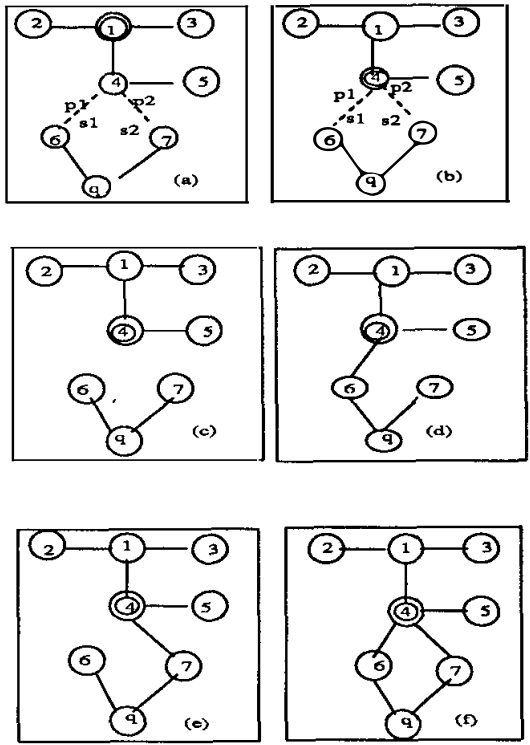

Figure 3: An example for illustrating configuration and transitions

A controlled transition sequence $T_{ij} = t_{ij}^1, ..., t_{ij}^k, k \geq 1$ is said to be *a generic transition* from a controlled configuration $i$ to a configuration $j$, if

$$j = trans(...trans(i, t_{ij}^1), ..., t_{ij}^k).$$

We use $gtrans(i, T)$ to denote the configuration resulting from *taking* generic transition $T$ in $i$. For the above case, we have $gtrans(i, T_{ij}) = j$. For a controlled transition $t$, let $c(t)$ denote the cost associated with $t$. The cost for a generic transition $T_{ij}$, denoted by $c(T_{ij})$, is the sum of the costs of the constituent transitions of $T_{ij}$.

Note that, for a given controlled configuration $i$ and a configuration $j$, there may exist zero, one or many generic transitions from $i$ to $j$. The *optimal generic transition* from $i$ to $j$ is the one with the least cost. A configuration $j$ is said to be a *generic successor* of a controlled configuration $i$ if there is a generic transition from $i$ to $j$, and configuration $j$ is either a terminal or is an uncontrolled configuration.

Intuitively, a configuration captures the characteristics of a situation, i.e. where the agent is now, where it wants to go and how much it knows about the environment. When an agent is in a controlled configuration, it can decide which edge to traverse next, and therefore, has control over the selection of the next (controlled) transition. Furthermore, if an agent takes a longer perspective on its navigation in a controlled configuration it can find that it will either enter an uncontrolled configuration first or enter a terminal first. Thus, we can consider the set of the generic transitions from $i$ to the generic successors of $i$ as the set of the possible actions the agent can take in configuration $i$. Obviously, if an agent wants to reach configuration $j$, it should take the optimal generic transition from $i$ to $j$. We use $GT(i)$ to denote the set of the optimal transitions from $i$ to the generic configurations of $i$.

When an agent is in an uncontrolled configuration, it faces one or more uncertain switches, and it can examine the states of the uncertain switches. When the new information is available to the agent, a new configuration is reached. Thus the agent has no control over which new configuration will be reached.

### 4.3 Navigation Procedure as a Markovian Decision Process

If we regard all the controlled configurations which may be encountered during a navigation collectively as a state space, all the generic transitions associated with each state as the possible actions which may be taken in that state, and the uncontrolled transitions as the state transitions, then, we can consider a navigation procedure as a Markovian decision process, and a navigation plan as a policy for such a decision process. We now give a formal account of this idea.

For a given U-graph $G = \langle V, E, S_u, pr, weight \rangle$, let $VSG(G)$ denote the following set:

$\{\langle V, E \cup S'_{on}, S'_u, pr, weight\rangle | S'_{on} \cup S'_u \subseteq S_u; S'_{on} \cap S'_u = \phi\}$.

In words, $VSG(G)$ denotes the collection of the variations of U-graph $G$ which are themselves U-graphs obtained by deleting some switches from $S_u$ and moving some of the remaining switches from $S_u$ to $E$.

For a given task of going from vertex $p$ to vertex $q$ in a U-graph $G$, let $IC$ be $\langle G, p, q \rangle$, called the *initial configuration*, and $I'$ be the set of all possible controlled configurations defined as:
$I' = \{\langle G', p', q \rangle | G' \in VSG(G), p' \in V, \langle G', p', q \rangle$ is a controlled configuration$\}$. Let $f$ denote a special state called *target state*. Let $I = I' \cup \{f\}$ be the state space.

In our current formulation, we consider $GT(i)$ as the set of the possible actions that can be taken in state $i$,

i.e. $K_i = GT(i)$. We assume further that there exists a special action $a_0$ which is the only action that could be taken in all the terminals. That is, for any state $i$ corresponding to a terminal configuration, $K_i$ is $\{a_0\}$ and taking action $a_0$ in state $i$ will result in the target state $f$.

For any state $i \in I$ and an action $a \in K_i$, the probability $q(i,j,a)$ is defined as:

$$q(i,j,a) = \begin{cases} 1 & \text{if } i \text{ is a terminal, } j = f \\ & \text{and } a = a_0; \\ 1 & \text{if } j = gtrans(i,a) \\ & \text{and } j \text{ is a terminal}; \\ p_t & \text{if } gtrans(i,a) \text{ is an uncontrolled} \\ & \text{configuration and} \\ & trans(gtrans(i,a),t) = j \\ & \text{for some transition } t \text{ in } gtrans(i,a) \\ 0 & \text{otherwise.} \end{cases}$$

where $p_t$ is the probability of transition $t$.

For any state $i \in I$, and an action $a \in K_i$, let $w(i,a)$ denote the cost for taking action $a$ in state $i$, which is defined as:

$$w(i,a) = \begin{cases} c(a) & \text{if } i \text{ is not a terminal}; \\ SD(G', n_i, q) & \text{if } a = a_0 \\ & \text{and } i \text{ is a good terminal}; \\ 0 & \text{otherwise.} \end{cases}$$

where $n_i$ is the current position of configuration $i$; $c(a)$ is the cost of generic transition $a$; $G'$ is the pessimistic induced graph of the U-graph of configuration $i$ and $SD(G', n_i, q)$ is the shortest distance from vertex $n_i$ to vertex $q$ in graph $G'$.

**Proposition 2** *For a given navigation task of going from vertex $p$ to vertex $q$ in a U-graph $G$, let $I$, $w$, $q$, and $K_i$ for each $i \in I$ be constructed as above, and let $K = \{K_i | i \in I\}$ and $M = \langle I, K, w, q \rangle$, then $M$ is a Markovian decision process.*

**Definition 3** A navigation plan *for a given navigation task is defined as a policy $R$ for the Markovian decision process corresponding to the task.*

**Definition 4** The expected cost *for a plan $R$ is given by $\lambda_R(IC)$, as defined in Section 4.1, where $IC$ is the initial configuration.*

**Definition 5** *For a given task,* the path planning problem *is to find a policy which minimizes $\lambda_R(IC)$.*

For a Markovian decision process $M$ derived from a navigation problem, we make the following important observation. That is: *"nature transitions" always lead to new states.* This means that, no matter what history the current state has, if an action (a generic transition) taken in this state will lead to an uncontrolled configuration which in turn will be led to a new controlled configuration (a new state), then this new state is different than any of the states in the history. The intuitive meaning is that whenever an agent reaches an uncontrolled configuration, it will find new information about the states of the uncertain switches, therefore, the resultant configuration must be different than any one encountered before. This implies that $RG_{i_0}(M)$, the representing graph of $M$, must be a directed acyclic graph (DAG).

### 4.4 Computing the Optimal Policy for a Navigation Problem

In the literature, the optimal first-passage problem was solved through successive approximations [Derman, 1970] or through computing the fixed point of a set of recursive equations [Denardo, 1982] [Eaton and Zadeh, 1962]. Both approaches involve iterations. The computational complexity of each iteration in these methods is $O(N * K)$, where $N = |I|$ is the size of state space and $K$ is the average number of possible actions that may be taken in a state. There is not a good upper bound for iteration times.

However, we can do much better for our case. As we mentioned earlier, the Markovian decision process $M$ derived from a navigation task with initial state $i_0$ can be represented by $RG_{i_0}(M)$ which is a DAG. A node in such a DAG represents a configuration which is reachable from the start state (configuration) $i_0$. The number of nodes in $RG_{i_0}(M)$ is no greater than $|I| + \sum_{i \in I} |K_i|$ even in the worst case. In the average case, the number of nodes is much smaller than the above bound, since, intuitively, many configurations may not be reachable from the initial configuration.

Now, let us consider the structure of an optimal policy $R*$. Obviously, we can assume $X(R*, f) = 0$. Since each state $i$ corresponding to a terminal configuration has only one action $a_0$, then $R*(i) = a_0$, thus, $X(R*, i) = w(i, a_0)$, for every state $i$ corresponding to a terminal.

For a state $i$ with possible action set $K_i$, let $J_{ia}$ denote the set of the possible next states after action $a$ is taken in state $i$, and $J_i = \cup_{a \in K_i} J_{ia}$ be the possible next states of state $i$. We have already known that:
$$X(R*, i) \leq X(R, i)$$
and
$$X(R, i) = w(i, a) + \sum_{j \in I} q(i, j, a) * X(R, j)$$
for any policy $R$ and all $i \in I$, where $a = R(i)$. Therefore, for a state $i$, suppose we already know $X(R*, j)$ for all $j \in J_i$, and suppose $a* = R*(i)$, then, $w(i, a*) + \sum_j q(i, j, a*) * X(R*, j)$ must equal $\min_{a \in K} \{w(i, a) + \sum_j q(i, j, a) * X(R*, j)\}$. Consequently, we have:
$$X(R*, i) = \min_{a \in K_i} \{w(i, a) + \sum_j q(i, j, a) * X(R*, j)\}.$$
(4)

This means that the value of $X(R*, i)$ can be computed easily if the values of $X(R*, j)$ are known for those $j \in J_i$.

Based on the above discussion, we obtain the following algorithm for computing $R*$ and $X(R*, i)$ for every $i$ which is reachable from $i_0$ in $RG_{i_0}(M)$.

**Algorithm A1**

Input: $RG_{i_0}(M)$;

Output: $RG_{i_0}(M, R*)$ and $X(R*, -)$ defined on the state nodes in $RG_{i_0}(M)$.

1. (Initialization) For each state $i$ corresponding to a terminal configuration, set $R*(i) = a_0$ and $X(R*, i) = w(i, a_0)$.
2. If $X(R*, i_0)$ has been computed, stop, otherwise, go to the next step.
3. For each state-node $i$, if $X(R*, i)$ has not been computed and for all $j \in J_i$, $X(R*, j)$ has been computed, compute $R*(i)$ and $X(R*, i)$ according to equation (4); cut off all the "non-optimal arcs" incident from $i$.
4. Go to step 2.

$RG_{i_0}(M, R*)$ can be obtained from the resultant graph by deleting all the nodes which are not reachable from $i_0$. Obviously, the complexity of the above algorithm is linear in the size of $RG_{i_0}(M)$. However, it should be noted that the size of $RG_{i_0}(M)$ can be exponential in the number of uncertain switches in a given U-graph.

## 5 RELATED WORK

At the application aspect, our work is related to the large body of research on spatial knowledge representation, path planning and navigation in AI and Robotics community. At the algorithmic aspect, our work is most closely related to some theoretical studies [Papadimitriou and Yannakakis, 1989] [Bar-Noy and Schieber, 1991].

**Related work in AI:** A number of approaches to path planning and navigation in uncertain environments have been proposed and many navigation systems for mobile robots have been built (e.g. [Arkin, 1989]; [Crowley, 1985]). However, in most of these systems, attention is primarily focused on the problem of how to make a robot capable of moving around in relatively small environments. Some notable exceptions are Kuipers' TOUR model [Kuipers, 1978], Kuipers and Levitt's COGNITIVE MAP [Kuipers and Levitt, 1988] for the representation of spatial knowledge of large scale environments, and Levitt and Lawton's qualitative approach [Levitt and Lawton, 1990] to navigation in large scale environments.

With respect to the express power, our U-graph model can express only the topological aspect of TOUR model in addition to a kind of uncertainty. While TOUR model supports both "highway oriented" navigation and "cross country" navigation, the U-graph supports only the former.

Dean et al [Dean et al., 1990] described a navigation system that also makes use of utility theory in navigation planning. However, their stress was on how to coordinate task achieving activities and map building activities so that a group of navigation tasks in an uncertain environment can be efficiently accomplished. In contrast, our planner is primarily concerned with the problem of finding the best way to accomplish a given navigation task.

A major difference between our work and the other related work (e.g. [Levitt and Lawton, 1990]; [Dean et al., 1990]) lies in the fact that the outcome of our path planning algorithm is not just a simple path, but is a comprehensive navigation plan which is highly conditional to the future states of the environment. In fact, the navigation plan generated by our path planner is somewhat similar to Schoppers' Universal Plan [Schoppers, 1989].

**Some related theoretical studies:** In [Papadimitriou and Yannakakis, 1989] Papadimitriou and Yannakakis first named the problem of travel under uncertainty as the *Canadian Traveller Problem* (CTP). In their formulation, a traveller is given an unreliable graph (map) whose edges may disappear. They also assume that the traveller cannot know whether an edge is actually there unless he/she reaches an adjacent vertex of the edge, and the status of an edge will not change after being revealed. The problem is to devise a travel strategy that results in an optimal travel (according to some predefined measure) from one vertex to another.

There is an obvious difference between our U-graphs and the uncertain graphs of the CTP with respect to environment modeling. In CTP, it is assumed that all the edges in a graph may disappear, and no information is known about the likelihood of the presence of any edge. However, in our U-graphs, we can make an explicit distinction between the "certain edges" and the "uncertain switches". Furthermore, by restricting the uncertain switches to a comparatively small number, we can reasonably assume that the prior probabilities of those uncertain switches can be obtained.

Besides the difference with respect to environment modeling, Papadimitriou and Yannakakis defined their optimality criteria in terms of *competitive ratio*. The competitive ratio is used in the literature to measure the quality of on-line algorithms [Mannasse et al., 1988]. Papadimitriou and Yannakakis showed that devising an on-line travel strategy with a bounded competitive ratio is PSPACE-complete. They also showed that the problem of devising a strategy that minimizes the expected ratio, provided that each edge in a graph has a presence probability, remains hard (#P-hard and solvable in polynomial space).

Bar-Noy and Schieber studied several interesting variations of the CTP [Bar-Noy and Schieber, 1991]. One variation is the *k-Canadian Traveller Problem*, which is a CTP with $k$ as the upper bound on the number of the blocked roads (disappeared edges). They gave a recursive algorithm to compute a travel strategy that guarantees the shortest worst-case travel time. The complexity of the algorithm is polynomial for any given constant $k$. They also proved that the problem is PSPACE-complete when $k$ is non-constant. Another variation Bar-Noy and Schieber studied is the *Stochastic Recoverable Canadian Traveller Problem*. In this problem, it is assumed that blocked roads can be re-opened in certain time. They presented a polynomial algorithm for devising a travel strategy that minimizes the expected travel time, under the assumption that for any edge $e$ in a given graph, the *recover time* of $e$ is less than the weight of any edge adjacent to it in the graph. Unfortunately, it is yet unclear how to relax this unrealistic assumption.

## 6 CONCLUSION AND FUTURE WORK

In this paper, we present U-graphs as a tool for modeling uncertain environments and formulate a U-graph based navigation problem as a Markovian decision process. This formulation contributes to a better understanding on the problem we try to solve. In addition, we present an optimality criterion for navigation plans and develop an algorithm for computing the optimal navigation plan *wrt* the criterion.

Future work can be carried out in several possible directions. First, we will study some other reasonable optimality criteria for path planning with uncertainty. For example, the optimality criteria could be quite different from the one discussed in this paper if we assume that an agent is subject to a cost limit. Second, we hope that we can make use of the hierarchical property of a given U-graph to cope with the complexity of the path planning with uncertainty. Third, we plan to derive some approximation algorithms having lower complexity, yet being able to compute reasonably good sub-optimal navigation plans for a given problem. Another kind of interesting future work is to incorporate our U-graph model into more general frameworks such as Kuipers' TOUR model [Kuipers, 1978] [Kuipers and Levitt, 1988].

### Acknowledgements

We would like to thank Ying Zhang and Andrew Csinger for their valuable comments on this paper. The work presented in this paper was supported in part by NSERC grant under operation number OG-POO44121.